\documentclass[letterpaper]{article} 
\usepackage{aaai2026}  
\usepackage{times}  
\usepackage{helvet}  
\usepackage{courier}  
\usepackage[hyphens]{url}  
\usepackage{graphicx} 
\usepackage{natbib}  
\usepackage{caption} 
\usepackage{algorithm}
\usepackage{algorithmic}

\usepackage{newfloat}
\usepackage{listings}
\DeclareCaptionStyle{ruled}{labelfont=normalfont,labelsep=colon,strut=off} 
\floatstyle{ruled}
\newfloat{listing}{tb}{lst}{}
\floatname{listing}{Listing}

\usepackage{amsmath}
\usepackage{amsthm}
\usepackage{amssymb}

\theoremstyle{definition}
\newtheorem{definition}{Definition}

\usepackage{tikz}
\usetikzlibrary{positioning,arrows.meta}

\title{Multiagent Reinforcement Learning for Liquidity Games}

\author{
  Alicia Vidler\textsuperscript{\rm 1},
  Gal A. Kaminka\textsuperscript{\rm 1}
}

\affiliations{
  \textsuperscript{\rm 1}Bar-Ilan University\\
  aliciavidler@gmail.com, galk@cs.biu.ac.il
}

\begin{document}
\sloppy
\maketitle

\begin{abstract}

Making use of swarm methods in financial market modeling of liquidity, and techniques from financial analysis in swarm analysis, holds the potential to advance both research areas. In swarm research, the use of game theory methods holds the promise of explaining observed phenomena of collective utility adherence with rational self-interested swarm participants. In financial markets, a better understanding of how independent financial agents may self-organize for the betterment and stability of the marketplace would be a boon for market design researchers. This paper unifies Liquidity Games, where trader payoffs depend on aggregate liquidity within a trade, with Rational Swarms, where decentralized agents use difference rewards to align self-interested learning with global objectives. We offer a theoretical frameworks where we define a swarm of traders whose collective objective is market liquidity provision while maintaining agent independence. Using difference rewards within a Markov team games framework, we show that individual liquidity-maximizing behaviors contribute to overall market liquidity without requiring coordination or collusion. This Financial Swarm model provides a framework for modeling  rational, independent agents where they achieve both individual profitability and collective market efficiency in bilateral asset markets.

\end{abstract}









\maketitle

\section{Introduction}


Cleared trading volume, and the system’s ability to achieve it efficiently, is commonly referred to as \texttt{liquidity}. In modern markets, it is an \emph{emergent} and \emph{desired} property of decentralized interactions among heterogeneous agents, not a fixed attribute. It denotes the ease of trading an asset quickly and at low cost without materially moving its price \cite{Fabozzi}. To quote Kevin Warsh, a member of the U.S. Federal Reserve Board of Governors: ``Liquidity is confidence''.

Increasing market liquidity is a challenging problem.  \emph{Market clearing}---turning intent into executed trades---comes from myriad local choices (quotes, inventories, risk limits) that produce system-level effects (such as aggregate trading volumes). However, since agent coalition and coordination are typically illegal in real-world modern financial markets, these local choices are \textit{required} to be uncoordinated. The trading intents (and their execution) of independent learners (agents) must be based on their  strictly \emph{local} rewards, that have no clear link to market-wide objectives.

In this paper, we address the tension between the \textit{self-interested} decisions of the independent agents, and the desired \textit{maximal liquidity} of market-wide trading volumes.
Specifically, we unify \emph{Liquidity Games (LG)}, a game-theoretic model of liquidity formation~\cite{vidler2024non}, with a decentralized multiagent reinforcement learning method, \emph{rational swarms}~\cite{kaminka2025swarms}. Here, multiple financial agents (investors, banks, or market makers) act on local information. Their collective decisions determine system-level liquidity patterns. The decentralized learning framework allows independent learners to act in a way that increases liquidity, while remaining independently self-interested.


Concretely,  we cast the market system as a Markov team-game, i.e., a fully cooperative stochastic repeated game, where agents theoretically receive the same global (system-level) payoff resulting from their joint actions. we treat \textit{aggregate cleared volume} as a global measure of liquidity. This is a system-level performance measure of the quantity of orders that are actually \emph{executed} (i.e., matched and settled) within a period, as distinct from quoted or submitted volume \cite{BOE}.
Aggregated cleared volume is the desired property to be maximized.
Following \cite{kaminka2025swarms}, we address the challenge resulting from the fact that agents are unable to receive (observe) this joint payoff, but are only exposed to a local proxy of it.
Instead, each agent learns using a local difference reward (marginal contribution to liquidity), yielding an individual reward signal that is aligned with global liquidity. This allows fully independent learning with no orchestration or collusion.
%
%
%
%
%
%
%
We examine two hypotheses:

\begin{description}
  \item[\textbf{[H1]: Liquidity with difference rewards}.] In bilateral trading with random pairing, agents trained with \emph{difference rewards} will achieve higher \emph{aggregate liquidity} and \emph{individual trading quantities} than those using other local or global signals, or baselines without learning, thereby maximizing local utility and global utility without coordination.

  \item[\textbf{[H2]: Robustness under exact match}.] In \emph{exact-match} Liquidity Games (no partial fills), agents trained with \emph{difference rewards} will learn policies that achieve and maintain higher \emph{matching efficiency} and will
  reach a high trade success threshold earlier in training than agents using alternative reward designs.
\end{description}



    We examine these hypotheses across different \textit{clearing regimes}, which define how trade offers convert to executed transactions and thereby shape the global objective \(G\) (cleared volume and hence, liqudity). We explore two regimes: (i) \emph{partial fill}, allowing fractional execution (which we term \textit{MinFill} regime) and (ii) \emph{exact match}, requiring full quantity alignment between buyers and sellers following \cite{vidler2024non}.

We show using difference rewards provides advantages over alternative learning and non-learning methods. Anticipating our results, under a partial clearing rule (MinFill), difference reward agents achieve the highest total liquidity and clearing efficiency; under exact matching, they remain competitive, rapidly reaching \(\sim70\%\) trade success despite stricter constraints.

This paper is structured as follows. First, 
we explore the relevant literature. Then, 
we detail the methodology and framework to combine difference reward functions of distributed multiagent reinforcement learning and the game theoretic model of bond markets focused on liquidity.  In the following sections, we outline specific instances of the combined paradigm and discuss the results.

\section{Financial Markets as Multiagent Systems:\\ A Literature Review}
\label{LitRev}

Traditional economic theory has employed several mathematical frameworks to model agent interactions in financial markets, each offering distinct perspectives on collective behavior and market dynamics.
Financial markets can be modeled as multiagent environments (MAS) where heterogeneous participants, ranging from retail traders to institutional algorithms, interact under shared protocols and constraints \cite{vidler2024decodingotcgovernmentbond, Axtell, Epstein2010, hull2022options}. The degree of agent heterogeneity varies between market segments, shaping systemic behavior and stability \cite{vidler2024decodingotcgovernmentbond}.

We suggest that there are theoretical and structural analogies between agents in financial markets~\cite{Vidler2025Agentic}, and multiagent reinforcement learning by independent learners in swarms~\cite{kaminka2025swarms}.
These can provide a foundation for understanding distributed decision-making in economic contexts.


\paragraph{Regulated Interaction and Implicit Coordination.} Unlike engineered MAS, financial systems restrict explicit cooperation due to legal prohibitions on collusion and information sharing \cite{FedClcear, BOE}. Despite this, coordination can still emerge implicitly through indirect channels such as strategic signaling, herding, and reaction to shared market data, resembling the stigmergic interactions observed in swarm systems. These markets thus function as decentralized, swarm-like systems, where coordination arises without central control. This analogy motivates our view of financial markets as distributed systems that process information and adapt dynamically, which we expand on below.


\paragraph{Distributed Information Processing.} Financial markets aggregate private information through the distributed trading mechanism, where individual beliefs are expressed through price signals (equity markets) and available liquidity (bond markets). This resembles distributed sensing in swarms. In some markets, such as government bonds, this process affects liquidity \cite{Pinter2023, Pinter2024, Vidler2025Agentic}.  System-level feedback highlights the tension in the system: Markets (systems) can adapt through feedback, but this adaptive capacity works best in smooth, unobstructed environments. When faced with significant challenges or resistance, the system's ability to maintain stable operation while adapting becomes limited. In bond markets, this has translated into several systemic shocks in the past few years in the UK and US markets alone (\cite{Pinter2023}, \cite{Duffie2020}).  Our work, in part, is motivated by these challenges and the opportunity to link markets when global goals are crucial (i.e. functioning bond markets) but where individual agents are both selfish and unable to explicitly coordinate to maximize global outcomes. We introduce literature specifically relevant to bond market liquidity below.


\subsection{Market Liquidity: Bilateral Market Structure and Liquidity Provision in Government Bond Markets}

Liquidity in bilateral, over-the-counter (OTC) markets emerges from decentralized interactions between counterparties rather than from centralized order books. Many institutional settings, including government bonds, concentrate asset trading. For example, in the UK market for government bonds (the gilt market), bonds are issued and traded via a network of regulated dealers (GEMMs)\footnote{https://www.dmo.gov.uk/responsibilities/gilt-market/market-participants/}, with broadly comparable arrangements in Australia, the US, and Canada \cite{Cheshire2015}. While institutional details vary, large investors in such markets typically transact through these intermediaries (or their functional equivalents).  Large participants supply quotes and warehouse risk but face balance-sheet, capital, and regulatory constraints \cite{Pinter2024, Czech2022}.


Across bilateral OTC environments, transaction prices are often anchored to public benchmarks or reference rates rather than discovered solely within each local match. Liquidity constraints arise from limits on inventories, funding capacity, and the topology of trading relationships \cite{Duffie_1999, Fabozzi, Pinter2023}. As a result, liquidity is an emergent property of many decentralized decisions (such as inventory management and risk tolerance) rather than a centrally controlled input \cite{Fabozzi}. Supervisors and monitoring authorities \cite{BOE} observe cleared volume: the amount of bonds which convert into executed trades, a metric that is used to asses \textbf{market liquidity}.  In this paper we will refer to market liquidity and liquidity, interchangeably.


    Gaps remain in how liquidity can be modeled as a global outcome emerging from local agent behavior. Existing approaches capture either structural features of market microstructure or the strategic interactions of participants. We address this by contributing a model that integrates the Liquidity Game (which formalizes the constraints and incentives of bilateral OTC trading \cite{vidler2024non}) with decentralized multiagent learning via rational swarms \cite{kaminka2025swarms}. This unified framework allows agents to act independently using local rewards, yet collectively reproduce system-level liquidity through emergent clearing (trading of bonds).  We add to this further by allowing agents to "learn" how to trade larger volumes, which increased aggregate system liquidity.  It thereby connects micro-level decision processes with macro-level market outcomes, providing a new method for studying liquidity formation without centralized coordination and agent policy learning.

\subsection{Game Theory applied to Financial Multiagent Systems}

    \paragraph{Game Theory.}
    Game-theoretic approaches have been applied to financial markets, but most frameworks analyze \textbf{micro-level interactions} rather than holistic market behavior, limiting insight into the emergent properties of large-scale systems of independent trading agents, such as liquidity analysis. The central challenge is scaling game-theoretic insights to capture the complex interdependencies in multi participant markets.

    Against this backdrop, applications of heterogeneous and homogeneous multiagent systems span small and largescale settings, notably in game theory, negotiation, and fairness \cite{Wooldridge_2009, Axtell2022, Parkes2006, Shoham_2008, PrinciplesKraus}, with related lines in \cite{Fagiolo_2017, Lussange_2020, Vermeir_2015}. The foundations for agent reasoning and communication provide logic-based frameworks for cooperation \cite{PrinciplesKraus, KluglRoleEnviron, ReasoningJOHN}. Extensions to automated negotiation and decision making under uncertainty blend game theory and AI \cite{Baumeister2019}, with results on repeated interaction and ``playing the wrong game'' bounds \cite{burkov2014repeated, meir2018playingwronggamebounding}. Research on goal modeling within ABMs includes \cite{winikoff2002declarative} and simulations are found to be crucial for reproducing complex dynamics \cite{Bai_2020}. These ideas underpin autonomous bidding agents (including financial agents) and market-oriented MAS explored in \cite{WellmanBook, lin2014genius, shoham2008computer}.  We make use of these elements and extend LG to include multiagent learning abilities, with the goal that market participant (agents) learn the best strategy for trading to mazimize their own liquidity (selfishly) while benefiting overall from global maximized market liquidity.

    \paragraph{Markov Games and Market Memory.} Markov decision processes and stochastic (Markov) games have found widespread application in financial modeling due to their mathematical tractability and the assumption of memory-less state transitions \cite{Shapley1953StochasticGames, Littman1994MarkovGames}. The Markov property assumes that future market states depend only on current conditions, independent of historical trajectories. However, this assumption has faced increasing scrutiny, as empirical evidence points to long-memory and path-dependent behavior in returns, volatility, and liquidity (order flow) \cite{DingGrangerEngle1993LongMemory, LilloFarmer2004LongMemoryOrders, LilloFarmer2005TheoryLongMemory}. In particular, volatility dynamics exhibits rough (fractional) behavior that departs from Markovian models \cite{GatheralJaissonRosenbaum2018}.
    Similarly, trading systems comprise repeated games, where participants see value in learning features of the counterparties they trade with \cite{MASSA200399}.

Across these approaches, \texttt{a gap remains}: existing models do not reconcile local agent incentives (with or without learning) and the emergence of global liquidity as an outcome.


We address this gap by contributing a unified framework that combines the structural rigor of the Liquidity Game \cite{vidler2024non} with decentralized learning via rational swarms \cite{kaminka2025swarms}. We show agents can learn locally, while preserving legal non-coordination, yielding a scalable, adaptive model of how liquidity emerges endogenously in complex financial systems.


\subsection{Swarm Characteristics of Financial Markets}

    \subsection{Reward mechanisms and swarms}

    Reward mechanisms in multiagent financial systems span individual incentives and system level objectives \cite{wooldridge1995intelligent, shoham2008computer}. At the agent level, rewards naturally map to monetary gains from trading. At the collective level, rewards correspond to attributes of market qualities such as stability, liquidity, and sustained participation. In certain markets (such as bond markets), agents derive direct benefit from liquidity (facility to transact the bonds they wish, at the time they desire) rather than from directional price dynamics.

    This paper seeks to unify two aspects of financial and swarm research: we build a unified framework for swarm-scale market liquidity that marries LG's (payoffs tied to aggregate liquidity \cite{vidler2024non})
    
with Rational Swarms (self-interested agents aligned via difference rewards \cite{kaminka2025swarms, TumerDifference}) inside a Markov team-game formalism.
  We make explicit that the difference reward is the marginal contribution of the agent to a global liquidity objective, providing a transparent, incentive compatible signal that preserves the independence of the agent and avoids coordination or collusion. This shifts the focus from two-party transactions to emergent, large population dynamics, linking individual trading incentives to system-level properties such as overall market liquidity.


\section{Model and Method Description}\label{Model}

In this section we detail our theoretical model definition.  The following subsections specify the environment (random pairing, clearing rules), agent types, and learning signals. All are defined relative to  a bilateral trading market in which the team objective is \emph{aggregate liquidity} \(G\) (total cleared volume).

\subsection{Liquidity Games}

    Liquidity Games are introduced in \cite{vidler2024non} and are defined as a ``noncooperative, simultaneous move game between two agents (or players), modeling a bilateral market mechanism''. Specific to the game is the unique feature of endogenous payoff.

        \begin{definition}[Liquidity Game]
        A game instance is a tuple \( I = (N,B,U) \) defined as follows:
        \begin{itemize}
          \item \textbf{Players.} \(N=\{i,j\}\) is the set of players, with \(|N|=2\).

          \item \textbf{Bond balances.} Each player \(i\in N\) holds a private bond balance
          \[
            B_i \in \mathbb{Z} \quad \text{with} \quad x \le B_i \le y \,,
          \]
          where \(x,y\in\mathbb{Z}\) are public market parameters. Typically \(x=0\) and \(y=c\), where \(c\) is the total issuance of the given bond (publicly available).

          \item \textbf{Actions.} The set of permitted actions for player \(i\) is
          \[
            A_i=\{0,1,\dots,|B_i|\}\,,
          \]
          and an action \(a_i\in A_i\) represents the number of bonds played by player \(i\) in a round. Equivalently,
          \[
            0 \le a_i \le |B_i| \, .
          \]

          \item \textbf{Utilities.} The utility (reward) function \(u_i:A_i\to\mathbb{Z}\) for player \(i\) is
          \[
            u_i(a_i;B_i)=|a_i| \quad \text{for all } a_i\in A_i .
          \]
        \end{itemize}
        \label{Def:LG}
        \end{definition}

\subsection{Game motivation: Repeated stage game}
    We model the process as a repeated (one-shot) stage game, where each episode the game state resets and multiple bilateral agent games occur each episode such that each agent has the opportunity to trade one time only per episode.  Within the stage game between two agents, simultaneously without the use of prices, each player adopts a (potentially) heterogeneous strategy by privately selecting a parcel $a_i \in B_i$ with $0 \le |a_i| \le |B_i|$ to trade, aiming to move their bond balance toward, but not below, zero while concealing true inventory to preserve informational advantages and satisfy regulatory norms. Choices may be formed on beliefs about the counterparty and agents reflect human players who typically seek full clearance in a single move.  Two game regimes are utilised: trades execute only if the submitted quantities are mutually acceptable and the  ``Minfill'' regime where the minumum is traded. Whilst outside the scope of this work, we note that these player preferences can be incorporated in more complex and richer game settings (such as penalty methods) and we leave this to future work.

    In the classic game of \cite{vidler2024non} payoff and welfare for a one-shot, simultaneous variant has one player, a seller with $B_i>0$ and a buyer with $B_j<0$ each choosing a non-negative action $a_k\in\mathbb{Z}_{\ge 0}$ with $a_k\le |B_k|$. Trade clears only on an exact positive match: $q=a_i$ if $a_i=a_j>0$, else $q=0$. The payoff for each player is $q$. This captures the shared goal of moving balances toward zero; any mismatch (or a zero submission) leaves positions unchanged.

Agents can belong to one of two mutually exclusive groups i.e. large or small firm size.

    Let there be a finite set of agents $N = \{1, \dots, n\}$.
    Each agent $i \in N$ belongs to one of two mutually exclusive types:
        \[
        \theta_i \in \Theta = \{\text{large}, \text{small}\}.
        \]

    The type $\theta_i$ represents the firm size of agent $i$.
    The size of the firm affects the mean of the bond pay-off distribution in any trade, but not its tail behaviour.

    We extend LG's in two ways to increase the richness of their applicability:

\subsubsection{Clearing Rules:}
    A trade is said to have occurred at the lower of $|a_i|$ and $|a_j|$ such that the minimum liquidity would be transacted between parties. We call this the "MinFill" variant. This then naturally extends the concept of the difference reward to be the value differential between what each agent had wished to trade, and what was traded. Formally

    We see that in Definition \ref{Def:LG} there is the possibility for at least two policies to exist in the market:
    \begin{enumerate}
        \item \textbf{LG-Exact (baseline).} Trade occurs iff $a_i=a_j>0$. The quantity traded is $q=a_i$, and $u_i(q)=q$.
        \item \textbf{LG-MinFill (extension).} Trade occurs whenever $a_i>0$ and $a_j>0$, with traded quantity $q=\min(a_i,a_j)$. The payoff is $u_i(q)=q$.
    \end{enumerate}

    Unless stated otherwise, the equilibrium statements for the baseline refer to LG-Exact, and the learning experiments marked ``MinFill'' implement the second rule. We note that in LG-Exact, every diagonal profile $(a,a)$ with $a\in\{1,\dots,\min(|B_i|,|B_j|)\}$ is a pure Nash equilibrium (and $(0,0)$ is also an equilibrium).
    In LG-MinFill, the unique pure Nash equilibrium is $a_i^\star=a_j^\star=\min(|B_i|,|B_j|)$.

    In extending LG's to two regimes we generalize beyond strict constraints.

 \subsection{Learning Agents}
    We extend the game to a repeated stage game, where initial agent balances are reset each episode, but agents can learn what is the optimal amount of bonds to try to trade.  Namely, agents seek to learn the amount of bond to offer that maximizes the trade amount.  The learning needs to take as context the probability what sort of agent is on the other side of the trade (which impacts the possible trade quantity), whilst seeking to maximize the total liquidity generated in a trade.



We adopt \emph{difference reward} of \cite{TumerDifference} as each agent's reward signal. We build extensively on work by \cite{kaminka2025swarms} which demonstrates the power of rational decentralized swarm agents to provide for environment level benefits.  For agent $i$ with joint outcome $\mathbf{z}$ and counterfactual $\mathbf{z}^{-i}$ (the system without $i$), the difference reward measures \emph{the agent's marginal contribution} to the global objective $G$ and is defined as:

\begin{equation}
	D_i \;\equiv\; G(\mathbf{z}) - G(\mathbf{z}^{-i})
	\label{eq:diff}
\end{equation}

In our setting, $G$ is a liquidity metric, so $D_i$ quantifies how much agent $i$ improves market liquidity while preserving agent independence and avoiding explicit coordination. The difference reward measures the marginal contribution of an agent's presence to the overall system performance.  Such marginal contribution can be quantified through metrics such as total trading volume for example.  In this way, applying difference rewards to bilateral trading environments (such as bonds as one example \cite{vidler2024non} would allow a framework to evaluate the nonlinear impact of agent inclusion versus exclusion, extending beyond simple proportional contributions to capture emergent effects arising from agent interactions. We propose making use of this approach to understand market liquidity.

This pairing of Liquidity Games (endogenous \(G\)) with Rational Swarms (difference rewards) aligns rational learning individually with the market-level goal while avoiding explicit coordination. We note
that this ``marginal contribution'' view is conceptually analogous to familiar ideas of risk attribution (e.g., marginal or incremental contributions to portfolio risk) \cite{Hull2018RMFI}.

\section{Experiments with a Unified Financial Swarms model}
   In this section we detail our programmatic model testing environment.  The code base implements a multiagent simulation environment to investigate coordination and liquidity provision in a simplified market setting. We use the simulation environment to explore the impact of different clearing rules and reward mechanisms on agent behavior and aggregate market outcomes.

    \subsection{Simulation environment}

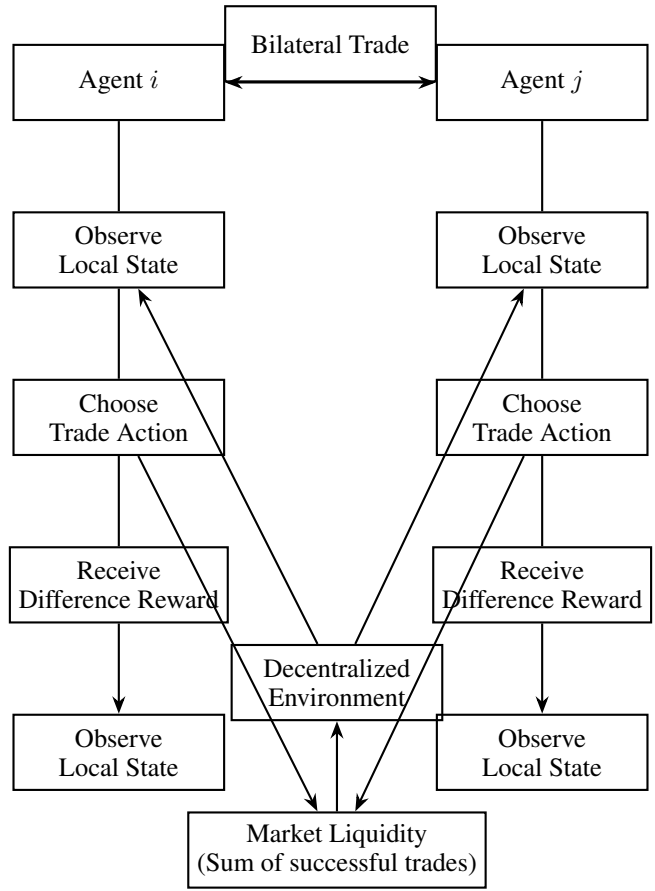
\begin{figure}[ht]
\centering
\begin{tikzpicture}[
  node distance=1.2cm and 2.8cm,
  every node/.style={draw, minimum width=2.8cm, minimum height=1cm, align=center},
  every path/.style={>=Stealth, ->, thick}
]

\node (agenti) {Agent \(i\)};
\node[below=of agenti] (obs1) {Observe\\Local State};
\node[below=of obs1] (choose1) {Choose\\Trade Action};
\node[below=of choose1] (reward1) {Receive\\Difference Reward};
\node[below=of reward1] (loop1) {Observe\\Local State};

\node[right=of agenti] (agentj) {Agent \(j\)};
\node[below=of agentj] (obs2) {Observe\\Local State};
\node[below=of obs2] (choose2) {Choose\\Trade Action};
\node[below=of choose2] (reward2) {Receive\\Difference Reward};
\node[below=of reward2] (loop2) {Observe\\Local State};

\node[below=2.5cm of reward1, xshift=2.9cm] (marketliq) {Market Liquidity\\(Sum of successful trades)};
\node[above=of marketliq] (env) {Decentralized\\Environment};

\draw (agenti) -> (obs1) -> (choose1) -> (reward1) -> (loop1);

\draw (agentj) -> (obs2) -> (choose2) -> (reward2) -> (loop2);

\draw[<->] (agenti.east) -- node[above]{Bilateral Trade} (agentj.west);

\draw (choose1) -- (marketliq);
\draw (choose2) -- (marketliq);
\draw (marketliq) -- (env);
\draw (env) -- (obs1);
\draw (env) -- (obs2);

\end{tikzpicture}
\caption{Schematic of the Liquidity Swarm model. Agents \(i\) and \(j\) interact via bilateral trades. Actions affect global liquidity, which feeds into local reward signals via difference rewards.}

\label{fig:liquidity-swarm}
\end{figure}

    The core of the simulation is the \texttt{LiquidityGameEnv} class, which models a market with a fixed number of agents (\texttt{n\_agents}). In each episode (time step), the agents are randomly paired. Within each pair, agents simultaneously announce an amount of liquidity they are willing to offer for trade to their compatriot.  Such details are never broadcast to other agents. The actual amount traded depends on the chosen clearing rule. After trades occur, agents' balances are updated, and a new episode begins with agents having potentially altered balances. Agents are designated as \textit{small} or \textit{large} based on their initial balance ranges (\texttt{max\_balance\_small} and \texttt{max\_balance\_large}), allowing for the study of asymmetric liquidity endowments. See Figure \ref{fig:liquidity-swarm} for a schematic representation of the interaction model.

\paragraph{Simulation settings.}
We make use of estimated agent numbers within real life markets in \cite{Pinter2023} and set agents to 1300, with large and small participants set at 33\% and 66\% respectively (drawing approximate values from \cite{vidler2024non}. The analysis is run on 10,000 episodes. The initial balance caps for each agent in each episode are set to small \(\le 10\) units; large \(\le 40\) units, where the agents initialize a balance as a random value between 1 and the cap at each episode, inline with the probability of being part of a large or small agent cohort. Our goal remains to test the ability for agents learning what amount of bonds to offer to trade such that they maximize their traded quantities and liquidity of the environment as a whole.

\subsection{Learning Agents (Tabular Q-Learning)}\label{learning}
We make use of simple Q learning agents.  However, the simulation compares the performance of learning agents employing different reward mechanisms against non-learning baseline agents.  All learning agents share the same tabular Q-learning core and differ only in the \emph{reward signal} (difference, local, or global).

\paragraph{Q-table.}
State \(s\) is the agent’s current balance; the action set is \(A(s)=\{1,\dots,s\}\) (offered amount). The Q-table has one row per discrete balance and one column per feasible action.

\paragraph{Update.}
After taking \(a\) in \(s\) and observing reward \(r\) and next state \(s'\),
\[
Q(s,a)\leftarrow Q(s,a)+\alpha\Big[r+\gamma\max_{a'}Q(s',a')-Q(s,a)\Big],
\]
with fixed \(\alpha=0.1\). The reward term \(r\) is instantiated by the chosen signal; no other hyperparameters differ across signals.

\paragraph{Action selection (epsilon-greedy).}
With probability \(\epsilon=0.2\) the agent explores by sampling uniformly from \(A(s)\); otherwise it exploits, choosing any \(a\in\arg\max_{a'}Q(s,a')\) (ties broken uniformly at random).






The learning agents are differentiated by the \textbf{reward mode} used to compute the signal they receive after each step.

\begin{enumerate}
    \item \textbf{Diff Reward:} Agents receive a difference reward, approximating their marginal contribution to the liquidity in the system. The reward for an agent is the quantity traded in their pair minus the quantity that would have been traded if they had offered zero (while their partner's offer remained the same). Under the \texttt{MinFill} rule, this simplifies to the trade quantity $q$.
    \item \textbf{Local Reward:} Agents receive a reward based on a local utility function: $R_i = 2q - a_i - \pi_{ij}$, where $q$ is the trade quantity, $a_i$ is the agent's offer, and $\pi_{ij}$ is a penalty for repeating a partner from the previous step, designed to explore adverse selection and set to a modest penalty of 0.1 units.

    \item \textbf{Global Reward:} Agents receive the total liquidity cleared across all pairs in the market during that step. This requires agents to learn to contribute to the global outcome without direct individual feedback on their impact.
\end{enumerate}

\subsubsection{Baseline Agents}

Two non-learning baseline strategies are implemented for comparison:

\begin{enumerate}
    \item \textbf{\texttt{random} Agent:} In each step, random agents offer a uniformly random amount between 1 and their current balance (inclusive).

    \item \textbf{\texttt{greedy} Agent:} In each step, greedy agents offer their entire current balance for trade. This strategy aims to maximize individual trade volume by being ``greedy''. A penalty is applied to the recorded liquidity for greedy agents under the \texttt{MinFill} rule if they offer more than their partner, reflecting a potential cost to over-offering and it set to \(20\%\) of the excess offered amount. In this way the agent is designed to be penalized for pure greed.
\end{enumerate}

\subsection{Metrics}

The simulation tracks several metrics per episode to evaluate the performance of different configurations: agent type (learning, non learning and different learning methods) and clearing rule (exact and MinFill).

\begin{itemize}
    \item \textbf{Cumulative Liquidity over episodes $G(z)$:} The sum of quantities traded across all pairs in a given episode (reported in Figure \ref{fig:cum_Liq} and \ref{fig:cum_liquidity}).
    \item \textbf{Convergence on clearing rates:} Looking at the smoothed percentage of each episodes initial balance that is traded between agents per episode. (see Figure \ref{fig:percent_cleared_exact}).

    \item \textbf{Hit Rate:} The fraction of paired agents that successfully complete a trade with $q>0$ (see Figure \ref{fig:Hit}).

\end{itemize}

By running experiments across the different clearing rules, reward modes, and baseline strategies, with configurable heterogeneity, the code base allows for a comprehensive analysis of factors influencing liquidity in this decentralized simulated market environment.

\section{Results}

\begin{figure}[h!]
  \centering
  \includegraphics[width=\columnwidth]{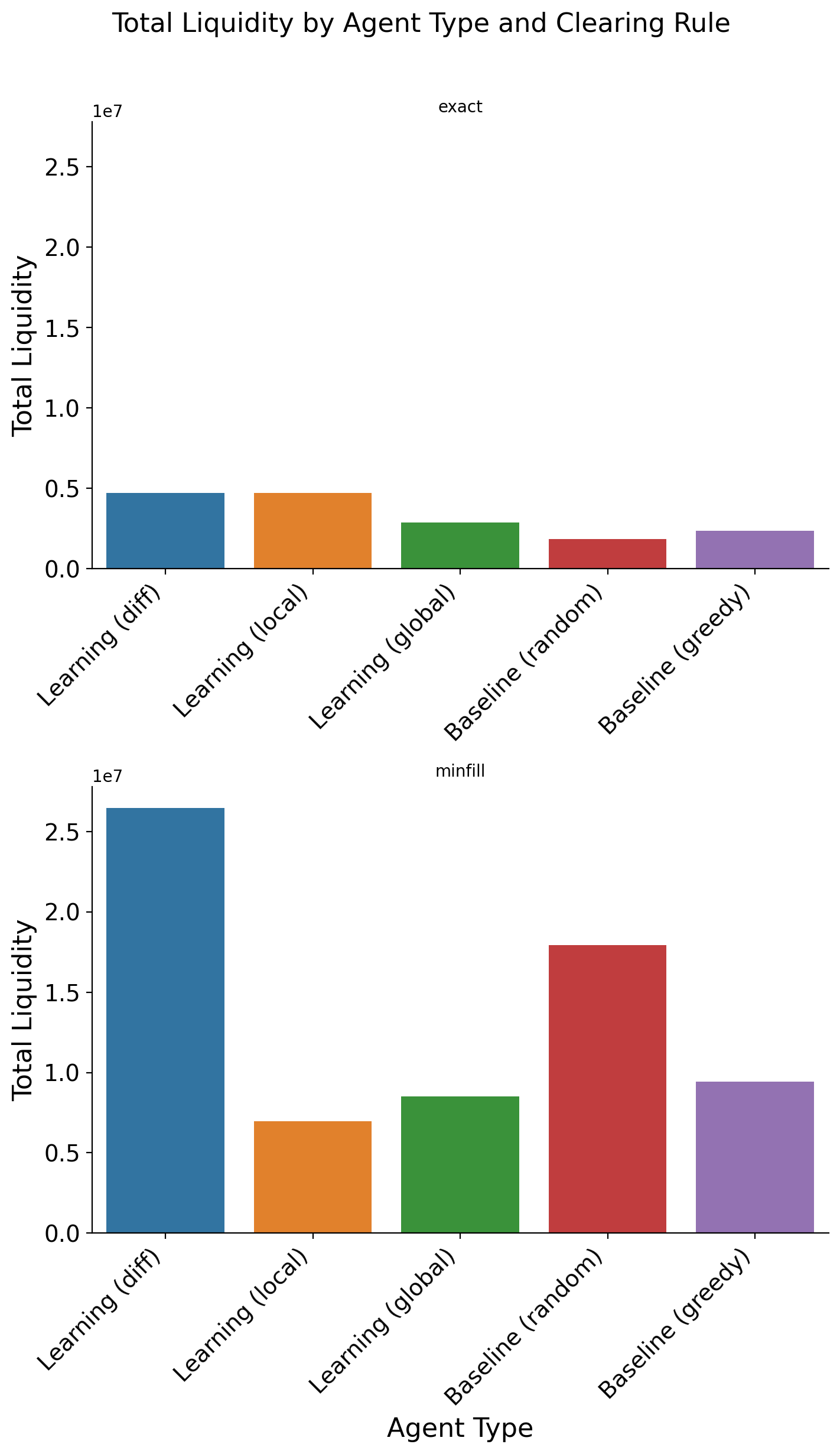}
  \caption{Learning agents surpassing baselines.}
  \label{fig:cum_liquidity}

\end{figure}

\begin{figure}[htbp] 
  \centering

  \includegraphics[width=\columnwidth]{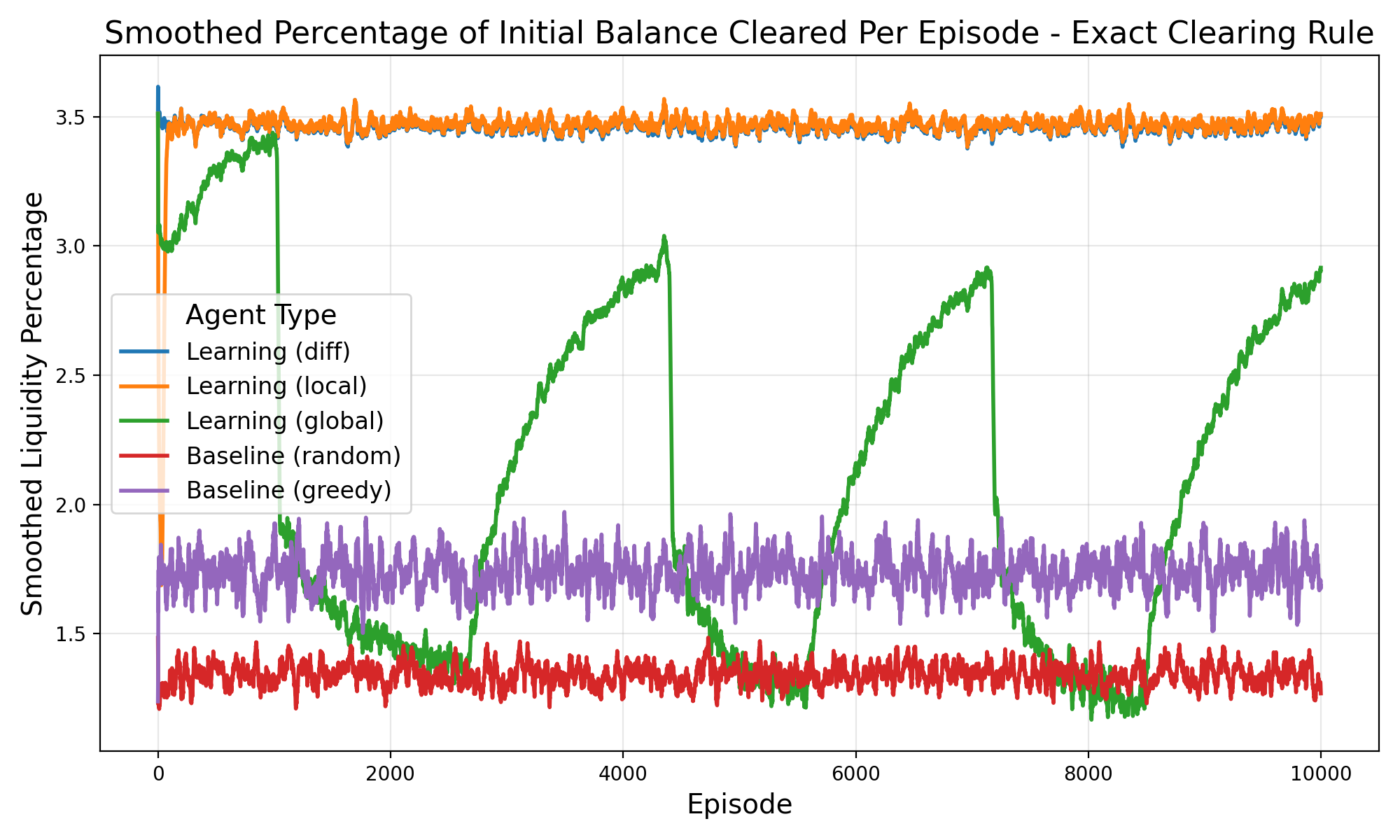}
  \caption{Clearing rates clearly converge early with learning, and difference
  functions in the exact rule being identical to local learning.}
  \label{fig:percent_cleared_exact}
\end{figure}

\begin{figure}[ht] 
  \centering
  \includegraphics[
    width=.45\textwidth,
    height=.4\textheight,
    keepaspectratio]{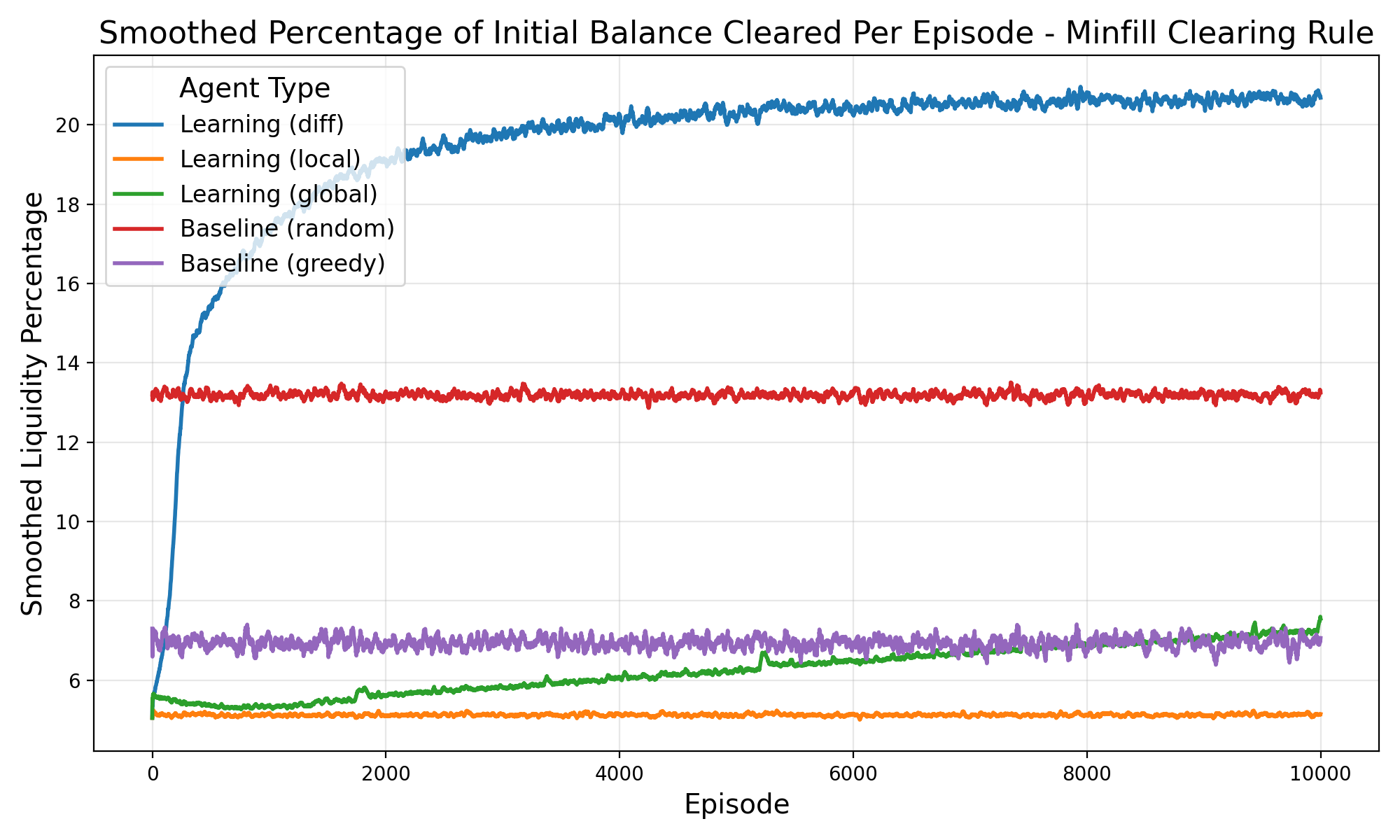}
  \caption{Clearing rates for Minfill learning.}
  \label{fig:percent_cleared_minfill}
\end{figure}

The experimental results demonstrate the effectiveness of the difference reward learning framework across different market clearing mechanisms (see Figure \ref{fig:cum_liquidity}). Under the `minfill` clearing rule, which permits partial order execution, the difference reward approach achieves the highest total liquidity cleared, outperforming both alternative learning methods ('Learning (global)' and 'Learning (local)') and baseline agents. This superiority is particularly pronounced compared to `exact` clearing, where the difference reward method maintains strong performance while other learning approaches converge to similar levels, although still exceeding the baseline benchmarks, as seen in Figure \ref{fig:percent_cleared_exact} and Figure \ref{fig:percent_cleared_minfill}.

\begin{figure}[ht]
  \centering
  \includegraphics[width=\columnwidth]{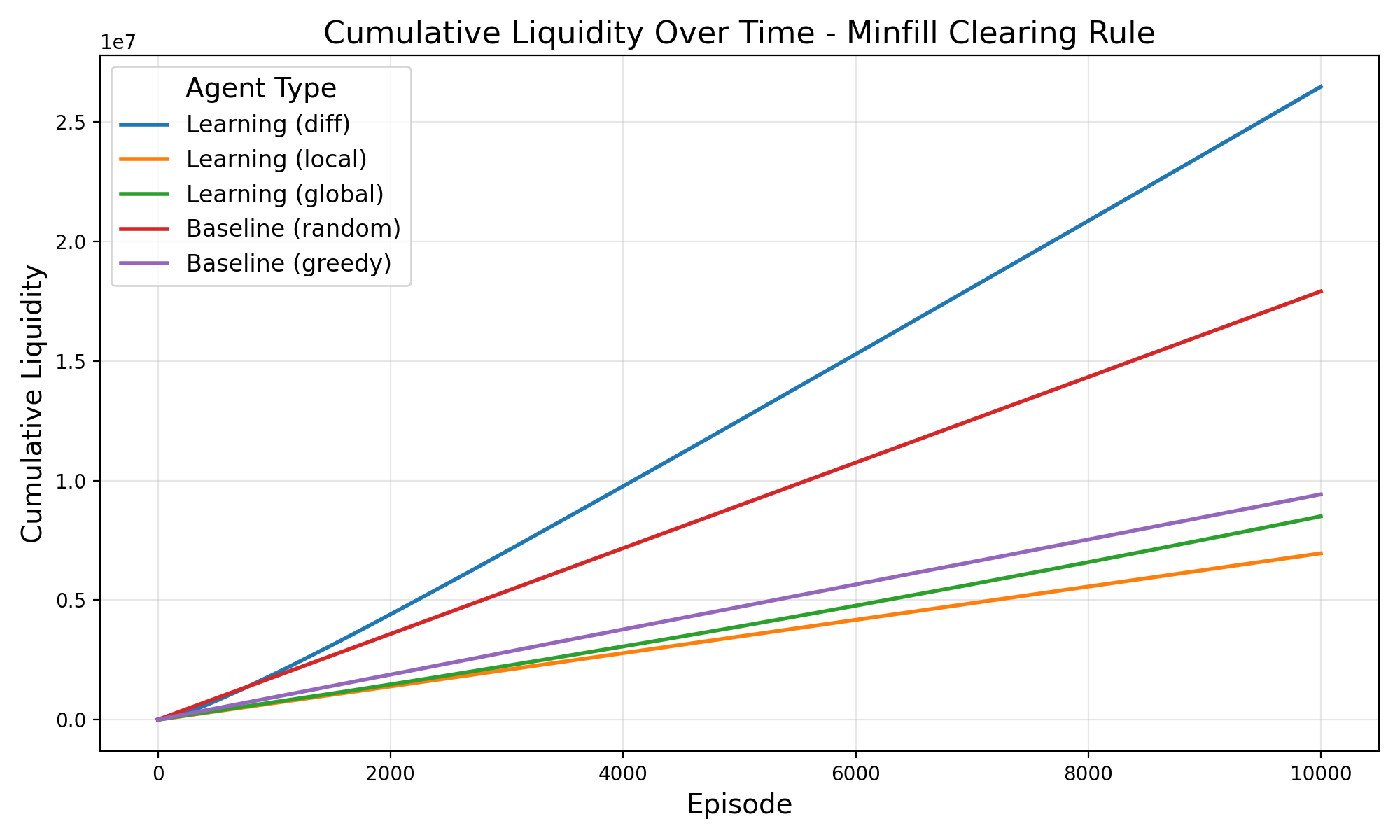}
  \caption{Cumulative liquidity through episodes for MinFill rules}
  \label{fig:cum_Liq}
\end{figure}

The cumulative liquidity trajectories reinforce the advantage of difference reward learning, especially under `minfill` where the framework's ability to attribute credit effectively translates into higher liquidity accumulation over episodes(Figure \ref{fig:cum_Liq}. In particular, the smoothed percentage of initial balance cleared per episode reveals that difference reward learning consistently achieves the highest clearing efficiency under `minfill`, demonstrating its capacity to maximize liquidity utilization relative to available resources.

\begin{figure}[htbp]
  \centering
  \includegraphics[width=\columnwidth]{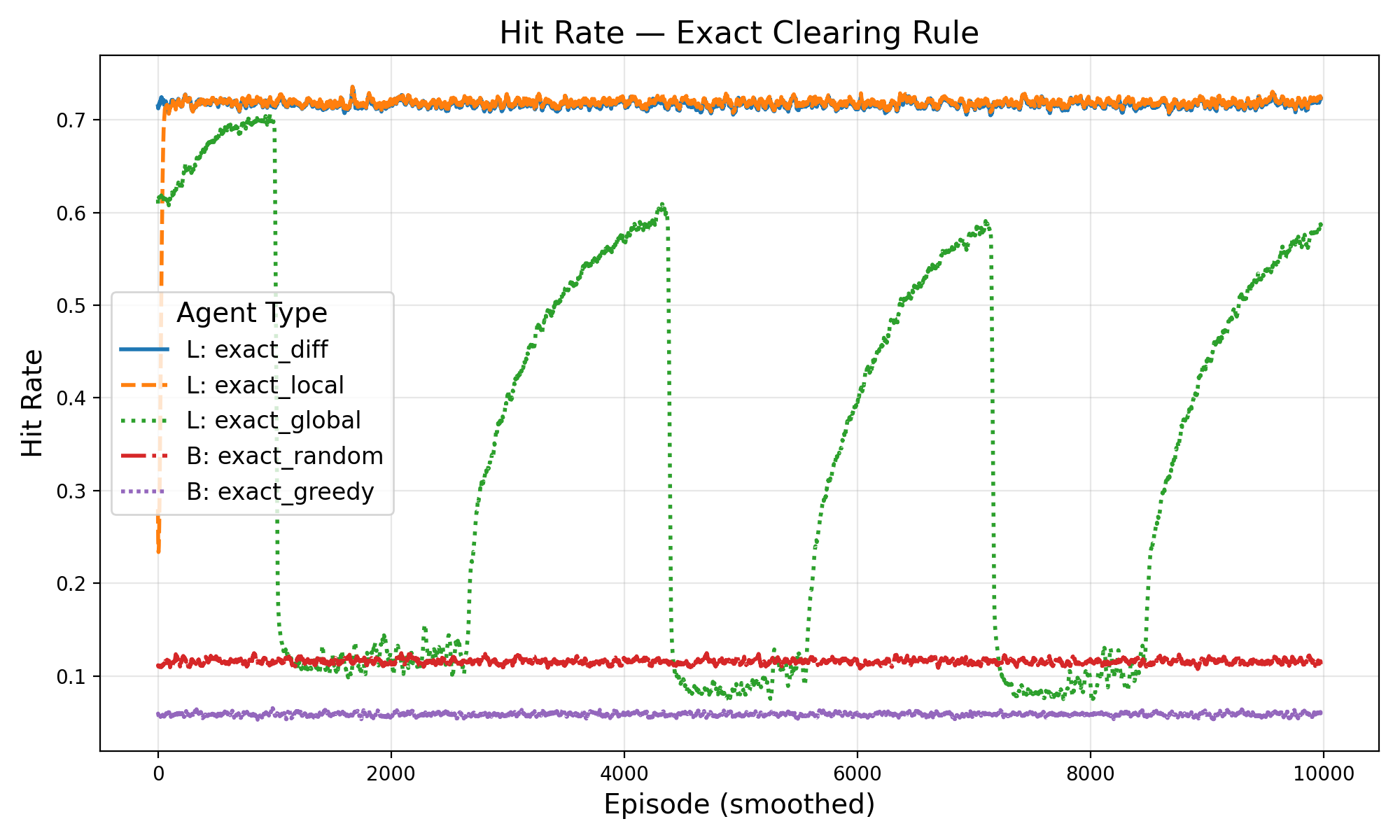}
  \caption{Hit ratios are high for learning algorithms, shown here for the exact fill rule (note: by construction, for MinFill they are 100\%). \textbf{Note:} Learning ($exact_diff$) and learning ($exact_local$) are almost identical results and graphs overwrite each other}
  \label{fig:Hit}
\end{figure}

As expected, regimes that permit partial clearing produced higher levels of liquidity, although learning agents in exact fill environments that used difference rewards were able to trade still successfully 70\% of the time despite constraints (Figure \ref{fig:Hit}.  These findings support difference rewards as a practical, incentive-compatible signal for liquidity provision when microstructure.

While the `exact` rule constrains overall performance and compresses differences between methods, the difference reward framework remains competitive, highlighting its robustness across varying market microstructures and its potential as a powerful approach for multiagent coordination in financial market simulations.

\section{Conclusion}
In this work, we show how to integrate Liquidity Games with decentralized difference-reward learning (Rational Swarms) to test whether self-interested agents can increase liquidity for themselves and the market. The results support both hypotheses. \textbf{H1:} Under the partial clearing rule (MinFill), the difference reward agents achieved the highest aggregate liquidity and per agent throughput, outperforming alternative local/global signals and non-learning baselines. \textbf{H2:} Even under exact-fill constraints, where performance gaps compress difference-reward agents, adapted quickly, reaching \(\sim70\%\) trade-success rates early in training and sustaining them thereafter. These findings indicate that difference rewards provide a practical and incentive-compatible signal for liquidity provision across microstructure regimes, with especially strong gains when partial clears are permitted. These results are particular helpful in the decentralized system where agent co-ordination is not acceptable.  Difference rewards implement marginal-contribution utilities: agents optimize private payoffs that track their contribution to total liquidity, promoting welfare-improving outcomes in bilateral markets.

In this paper, we demonstrate that by combining Liquidity Games with Rational Swarms, it is possible to design local incentives such that myopic best responses yield system level gains, turning the liquidity externality into an individually rational objective. This offers a practical template for scalable market design without the need for central coordinators or agent co-ordination.

\paragraph{Limitations and next steps.}
Our experiments use stylized dynamics (random pairing, reset endowments, no price formation), so external validity requires caution. Future work will (i) add persistent state and frictions (inventory, transaction costs), (ii) test richer clearing mechanisms and market impact models, and (iii) analyze sensitivity to agent heterogeneity and partial observability. Extending the analysis to real data or hybrid ABM–RL settings can further probe how difference rewards scale.

\bibliography{ref2025,Markov}

\end{document}